\documentclass[10pt,letterpaper]{article}
\usepackage[margin=1in]{geometry}
\usepackage{times}
\usepackage{natbib}
\usepackage{booktabs}
\usepackage{multirow}
\usepackage{graphicx}
\usepackage{amsmath,amssymb}
\usepackage{caption}
\usepackage{tabularx}
\usepackage{array}
\usepackage{subcaption}
\usepackage{hyperref}
\usepackage{url}

\title{GeoForward: Geometry-Guided Cross-View Alignment for Feedforward 3D Gaussian Splatting}
\author{Yiming Shu\textsuperscript{1}\thanks{Equal contribution.}, Peng Sun\textsuperscript{2}\footnotemark[1],
Lang Zhang\textsuperscript{2}\thanks{Project leader.}, Mingyue Feng\textsuperscript{2},\\
Yaonong Wang\textsuperscript{2}, Pei Liu\textsuperscript{3}\thanks{Corresponding author.},
Chen Sun\textsuperscript{1}, Jun Ma\textsuperscript{3}\\[0.7em]
\small \textsuperscript{1}The University of Hong Kong\\
\small \textsuperscript{2}Leapmotor\\
\small \textsuperscript{3}The Hong Kong University of Science and Technology (Guangzhou)}
\date{}
\hypersetup{
  pdftitle={GeoForward: Geometry-Guided Cross-View Alignment for Feedforward 3D Gaussian Splatting},
  pdfauthor={Yiming Shu, Peng Sun, Lang Zhang, Mingyue Feng, Yaonong Wang, Pei Liu, Chen Sun, Jun Ma}
}

\begin{document}

\maketitle
\vspace{-0.2cm}
\begin{figure}[h]
  \centering
\includegraphics[width=1.0\linewidth]{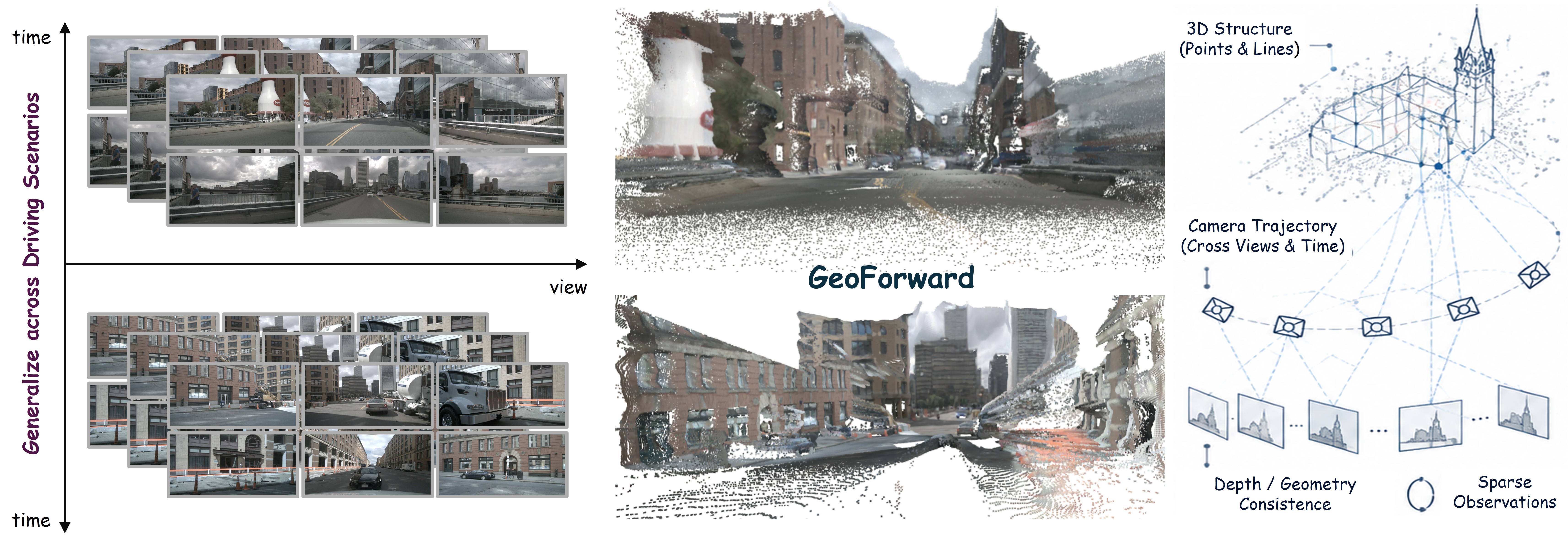}
\caption{\textbf{From diverse observations to a unified 3D scene.} Driving scenes are observed from multiple cameras and timestamps, spanning different scenarios and viewpoints. GeoForward transforms these heterogeneous observations into a unified Gaussian scene representation, enabling geometrically consistent reconstruction across views and time.}
  \label{fig:teaser}
\end{figure}

\begin{abstract}
Feedforward 3D reconstruction has largely followed an {appearance-driven paradigm}: multi-view images are encoded into visual representations, cross-view correspondence is implicitly inferred from learned features, and 3D primitives are subsequently predicted from these representations. While effective, this paradigm leaves a fundamental source of information underexploited: calibrated cameras and sparse metric measurements already provide explicit constraints on where the same physical point should appear across views. We propose a novel paradigm, \textbf{Geometry-Conditioned Feedforward Reconstruction}, in which metric geometry establishes cross-view correspondence before the 3D representation is formed, while learned visual features refine rather than discover this correspondence. Based on this paradigm, we introduce \textbf{GeoForward}. Sparse LiDAR points are unprojected into 3D and analytically projected across calibrated cameras to provide metric correspondence anchors. A bounded continuous local search then refines these anchors using visual features, and the resulting geometry-conditioned representation is fused with global multi-view context from VGGT. The shared representation drives separate metric-depth and Gaussian-attribute prediction, establishing a direct path from explicit geometry to both Gaussian centers and spatial attributes. A sparse 3D refinement module subsequently improves local attribute compatibility while preserving the geometrically determined centers. For temporal reconstruction, endpoint Gaussians are composed using vehicle masks and motion priors. On nuScenes, GeoForward achieves the lowest cross-camera depth-consistency error and Chamfer distance among the compared methods while maintaining competitive reconstruction quality. These results demonstrate that explicitly conditioning correspondence formation on metric geometry provides an effective alternative to purely appearance-driven feedforward 3D reconstruction.
\end{abstract}

\section{Introduction}
\label{sec:introduction}
\vspace{-0.05cm}
Feed-forward 3D reconstruction has recently emerged as a promising alternative to optimization-based scene reconstruction, enabling 3D representations to be predicted directly from multi-view observations without per-scene fitting~\citep{smyzanowicz2024splatter,charatan2024pixelsplat,chen2024mvsplat,tian2025drivingforward}. 
A central challenge in this setting is establishing reliable correspondence across views: observations from different cameras should describe the same physical surface and therefore produce geometrically compatible representations. Existing approaches typically address this challenge by learning cross-view relationships from visual representations, where image features implicitly encode correspondence cues before 3D geometry is decoded. This appearance-driven formulation has achieved impressive performance, but it leaves an important source of information underexploited. In calibrated autonomous-driving platforms equipped with sparse LiDAR, metric measurements already provide physical constraints on where a scene point observed in one camera should appear in another.

This motivates a fundamental question: \emph{can metric geometry directly participate in cross-view correspondence formation before 3D representation prediction?} 
Existing feed-forward reconstruction methods incorporate geometric information through different mechanisms, including learned depth prediction, explicit depth-hypothesis reasoning, or geometric foundation models that infer multi-view structure from images~\citep{charatan2024pixelsplat,chen2024mvsplat,wang2024dust3r,wang2025vggt}. These approaches demonstrate the power of learned visual representations for recovering geometry, but the correspondence underlying reconstruction remains primarily obtained through visual inference. In driving scenes, sparse LiDAR and calibrated cameras provide a complementary opportunity: instead of requiring a network to discover all possible cross-view associations from appearance, metric observations can provide a physically grounded initialization that narrows the correspondence search space.

We introduce \textbf{GeoForward}, a geometry-conditioned paradigm for feed-forward 3D reconstruction, as illustrated in Fig.~\ref{fig:teaser}. 
Different from appearance-driven correspondence learning, GeoForward first establishes metric cross-view correspondence using sparse LiDAR measurements and calibrated camera geometry. A LiDAR-supported point in a reference view is unprojected into 3D and projected into context views, providing an explicit correspondence anchor. Since sparse sensing, calibration errors, and discretization introduce residual uncertainty, a bounded continuous refinement module further adjusts the correspondence using local visual features. In this way, geometry constrains \emph{where} cross-view evidence should be associated, while learning determines \emph{how} the local correspondence should be refined.

The geometry-conditioned representation is then used for Gaussian scene reconstruction. GeoForward fuses the aligned features with global multi-view context from a frozen VGGT encoder~\citep{wang2025vggt} and jointly predicts metric depth and Gaussian attributes. The depth branch determines Gaussian centers through camera unprojection, while the attribute branch predicts spatial extent, opacity, and appearance conditioned on the geometry-aware representation. Beyond initial prediction, a sparse 3D neighborhood refinement module improves local attribute compatibility while preserving the predicted Gaussian centers. For dynamic driving scenes, endpoint Gaussian representations are further composed at arbitrary query times using vehicle masks and motion priors~\citep{yu2026recondrive,he2026dynamicvggt,song2026envision4d}.

We evaluate GeoForward on two large-scale driving-scene benchmarks, nuScenes~\cite{caesar2020nuscenes} and Waymo~\cite{sun2020scalability}. Experiments demonstrate that explicitly conditioning correspondence formation on metric geometry improves cross-view geometric consistency while maintaining competitive rendering quality. These results suggest a complementary direction for feed-forward reconstruction: rather than relying solely on learned appearance-based association, available metric information can directly participate in forming the 3D representation.
Our contributions are summarized as follows:
\begin{list}{\textbullet}{
\setlength{\leftmargin}{0pt}
\setlength{\rightmargin}{0pt}
\setlength{\itemindent}{0pt}
\setlength{\listparindent}{0pt}
\setlength{\labelsep}{0.5em}
}
\item We propose a \textbf{geometry-conditioned paradigm for feed-forward 3D reconstruction}, where sparse metric geometry constrains cross-view correspondence before 3D representation formation, reducing ambiguous visual matches and anchoring depth and Gaussian prediction in metric space.

\item We introduce \textbf{GeoForward}, which combines LiDAR-based correspondence initialization with bounded visual refinement and propagates geometry-conditioned features into Gaussian geometry and attribute prediction.

\item We introduce \textbf{sparse 3D attribute refinement} to improve local Gaussian consistency without modifying geometrically determined centers, and demonstrate improved cross-view reconstruction quality on \textbf{nuScenes and Waymo}.

\end{list}

\section{Related Work}

\subsection{Feed-Forward 3D Scene Reconstruction}

Recent advances in feed-forward reconstruction aim to predict 3D scene representations directly from multi-view observations, avoiding expensive per-scene optimization. DUSt3R learns pairwise point maps from image pairs, while VGGT extends this idea to jointly estimate depth, point maps, camera parameters, and point tracks from multiple views~\citep{wang2024dust3r,wang2025vggt}. InfiniDepth further explores continuous coordinate-query decoding for high-resolution geometric prediction~\citep{yu2026infinidepth}. 

Building on these geometric representations, recent works investigate feed-forward Gaussian reconstruction. MVSplat, AnySplat, and IDESplat predict Gaussian representations directly from images, improving reconstruction efficiency and scalability~\citep{chen2024mvsplat,jiang2025anysplat,long2026idesplat}. For autonomous driving scenarios, DrivingForward, DGGT, and StreetForward incorporate camera poses, temporal information, and dynamic scene modeling into feed-forward Gaussian reconstruction~\citep{tian2025drivingforward,chen2026dggt,yu2026streetforward}. DynamicVGGT and Envision4D further extend feed-forward reconstruction toward temporal prediction and future scene generation~\citep{he2026dynamicvggt,song2026envision4d}. Other approaches improve multi-view representation through 3D latent structures, uncertainty-aware canonical representations, or sparse 3D queries, including UniSplat, CanonicalGS, and PointForward~\citep{shi2026unisplat,xu2026learning,chi2026pointforward}. ReconDrive further combines VGGT representations with hybrid Gaussian prediction and velocity-based temporal composition for driving scenes~\citep{yu2026recondrive}. 

Although these methods achieve impressive reconstruction quality, cross-view consistency is mainly obtained through learned visual representations and implicit feature interaction. GeoForward complements this direction by introducing explicit metric constraints into the cross-view association process before Gaussian prediction.

\subsection{Metric-Guided Multi-View Geometric Reasoning}

Metric measurements provide an important source of geometric constraints for multi-view reconstruction and depth estimation. Previous works incorporate sparse geometric priors into multi-view stereo through sparse cost-volume construction or depth hypothesis refinement~\citep{qi2022sparse,poggi2022multi}. In autonomous driving, LiDAR measurements are particularly valuable due to their accurate metric information, and several approaches explore LiDAR-guided geometric prediction. DriveMVS uses sparse LiDAR observations to anchor depth hypotheses and incorporate geometric cues into feature fusion, while DrivingDepth leverages sparse LiDAR measurements to calibrate monocular depth estimation through metric correction~\citep{sun2026lidar,huang2026drivingdepth}. Related LiDAR-camera fusion methods further combine sparse depth measurements with image features for dense depth estimation~\citep{fu2019lidar,wei2023real}.

These approaches demonstrate the effectiveness of metric information for reducing geometric ambiguity. However, existing methods primarily introduce geometry as a constraint on depth estimation or feature enhancement, while the cross-view association underlying representation prediction is still largely determined by learned visual similarity. GeoForward explores a different use of metric information by treating sparse LiDAR measurements as geometric anchors for cross-view feature alignment. Rather than predicting correspondence purely from appearance or depth hypotheses, GeoForward establishes geometry-aware associations through calibrated projection and refines them using local visual evidence.

\begin{figure}[t]
  \centering
  \includegraphics[scale=0.195,trim=0 -2 0 0, clip]{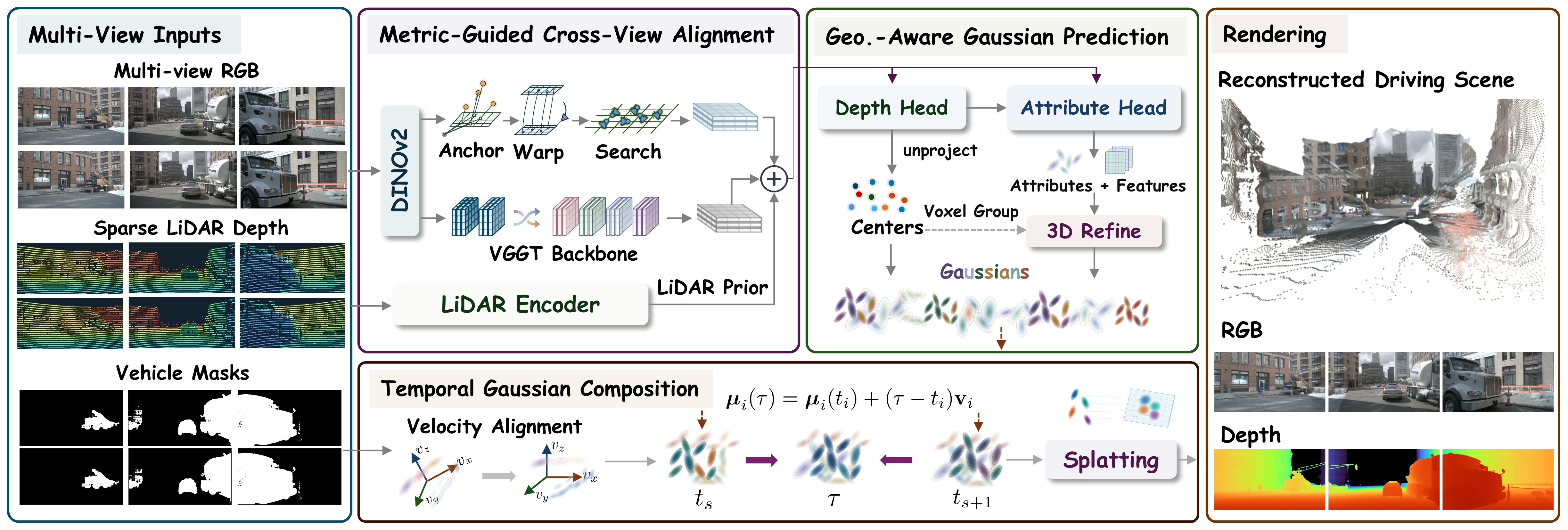}
\caption{\textbf{Overview of GeoForward.}
Given calibrated multi-view images and sparse LiDAR, GeoForward first establishes metric cross-camera correspondences through geometric projection and continuous local refinement. The aligned features are fused with global multi-view context and jointly condition metric depth and Gaussian-attribute prediction. A sparse 3D refinement module further improves local attribute consistency, while vehicle masks and motion annotations enable temporal Gaussian composition at arbitrary query times.}
  \label{fig:geoforward_overview}
\end{figure}

\section{Methodology}
\label{sec:geoforward_method}
\begingroup
\setlength{\abovedisplayskip}{5pt plus 2pt minus 1pt}
\setlength{\belowdisplayskip}{5pt plus 2pt minus 1pt}
\setlength{\abovedisplayshortskip}{3pt plus 1pt minus 1pt}
\setlength{\belowdisplayshortskip}{3pt plus 1pt minus 1pt}
\setlength{\parskip}{0pt}

\textbf{Overview.}
GeoForward reconstructs a Gaussian scene from calibrated surround-view images $I_{t,v}$ and sparse LiDAR depth $D^{\mathrm L}_{t,v}$ observed at two endpoint timestamps. Camera intrinsics $K_{t,v}$ and camera-to-reference poses $E_{t,v}=(R_{t,v},\mathbf{t}_{t,v})$ transform all observations into a common reference-ego frame. The key idea of GeoForward is to convert sparse metric measurements into explicit {cross-camera feature correspondence} before Gaussian prediction. As illustrated in Fig.~\ref{fig:geoforward_overview}, a LiDAR-supported pixel in a reference camera is first projected into a context camera using calibrated geometry, providing a metric correspondence initialization. A bounded learned offset then refines this correspondence using local visual evidence. The resulting aligned features are combined with global multi-view features from a frozen VGGT encoder~\citep{wang2025vggt} and shared by the depth and Gaussian-attribute prediction branches. Finally, a sparse 3D refinement module improves local attribute compatibility while preserving the predicted Gaussian centers, and endpoint Gaussians are composed at arbitrary query times for dynamic-scene rendering. Sparse LiDAR is used at both training and inference.

\subsection{Metric-Guided Cross-View Alignment}
\label{sec:geoforward_alignment}

The central component of GeoForward is metric-guided cross-view alignment, as illustrated in Fig.~\ref{fig:geoforward_alignment}. Rather than requiring a network to infer all cross-camera correspondence implicitly, we use sparse LiDAR measurements to provide an analytically determined correspondence and learn only a bounded local correction. This separates the problem into two complementary components: {metric initialization} from calibrated geometry and {visual refinement} from learned features.

We use a frozen VGGT encoder~\citep{wang2025vggt} to jointly process the endpoint images. Its global multi-view representation is retained to provide scene-level context, while intermediate DINOv2 features~\citep{oquab2023dinov2} are used for local correspondence refinement. Freezing VGGT isolates the contribution of explicit metric correspondence and avoids relying on additional end-to-end adaptation of the geometric foundation model.

\noindent\textbf{Metric correspondence initialization.}
For a reference view $A$, a sampled LiDAR depth at image location $\mathbf{p}$ defines a 3D anchor $\mathbf{X}_A = d_A^{\mathrm L}(\mathbf{p}) K_A^{-1} \widetilde{\mathbf{p}},$ where $\widetilde{\mathbf{p}}$ is the homogeneous image coordinate. Given the relative pose $E_B^{-1}E_A = \left( R_{B\leftarrow A}, \mathbf{t}_{B\leftarrow A} \right),$ the anchor is transformed into context view $B$ and projected onto its image plane:
\begin{equation}
\mathbf{g}_{AB}(\mathbf{p}) = \pi\left( K_B \left( R_{B\leftarrow A}\mathbf{X}_A + \mathbf{t}_{B\leftarrow A} \right) \right),
\label{eq:gf_warp}
\end{equation}
where $\pi(\cdot)$ denotes perspective division. We construct a support mask $m_{AB}$ according to LiDAR availability, positive projected depth, image validity, and context-view visibility.

For cross-time alignment, the same projection is applied using the calibrated poses under a static-scene assumption. This approximation is used only to establish the initial visual correspondence; object motion is handled separately during temporal Gaussian composition in Sec.~\ref{sec:geoforward_temporal}.

\begin{figure*}[t]
\centering
\includegraphics[scale=0.2,trim=0 -2 0 0,clip]{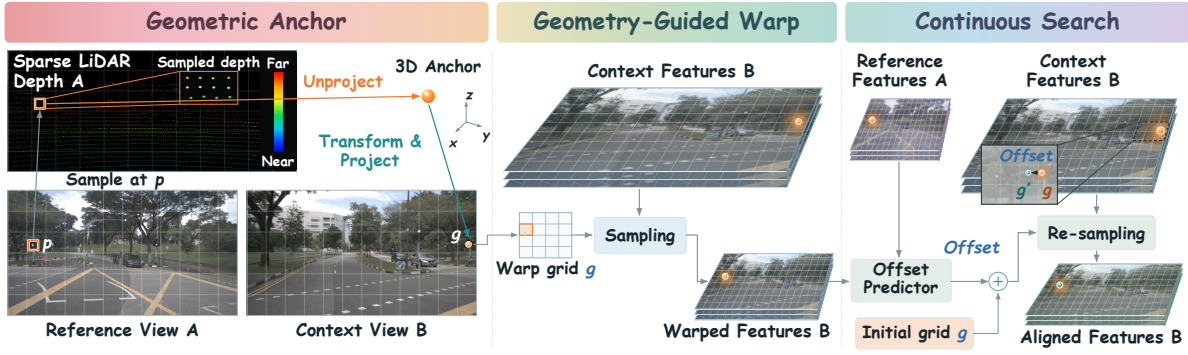}
\caption{\textbf{Metric-guided cross-view alignment.}
Sparse LiDAR provides a metric correspondence anchor by projecting a 3D point from the reference view into a context view using calibrated geometry. A bounded continuous offset then locally refines the projected location using visual features, after which the context feature is re-sampled at the refined coordinate.}
\label{fig:geoforward_alignment}
\end{figure*}

\noindent\textbf{Learned local refinement.}
The metric projection provides a strong initialization but does not guarantee pixel-level feature correspondence because of sparse sampling, calibration errors, discretization, and view-dependent appearance. We therefore refine the projected location using local feature evidence. Let $\mathcal{S}$ denote the feature-query operator, we obtain $\mathbf{f}_A = \mathcal{S}(F_A,\mathbf{p}),$ $\mathbf{f}^{0}_{AB} = \mathcal{S} \left( \operatorname{sg}(F_B), \mathbf{g}_{AB} \right),$ where $F_A$ and $F_B$ are intermediate feature maps and $\operatorname{sg}(\cdot)$ denotes the stop-gradient operator.

An offset predictor takes the local feature pair as input and predicts a bounded correction:
\[
\Delta\mathbf{g}_{AB} = r\tanh \left( \Phi \left( [\mathbf{f}_A,\mathbf{f}^{0}_{AB}] \right) \right),
\]
where $r$ specifies the maximum search radius in pixels. The refined context feature is then obtained by re-sampling the original context feature map:
\begin{equation}
\widehat{\mathbf{f}}_{AB} = \mathcal{S} \left( \operatorname{sg}(F_B), \mathbf{g}_{AB} + \Delta\mathbf{g}_{AB} \right).
\label{eq:gf_search}
\end{equation}
The bounded formulation restricts the learned refinement to a local neighborhood around the metric initialization rather than learning correspondence from scratch. Gradients propagate through the sampling coordinates while the sampled context feature values are detached.

For a reference view, refined features from supported context views are aggregated as $\overline{\mathbf{f}}_A = \frac{ \sum_B m_{AB} \widehat{\mathbf{f}}_{AB} }{ \sum_B m_{AB}+\epsilon }.$ When no valid context view is available, we set $\overline{\mathbf{f}}_A=\mathbf{0}$. A residual fusion layer incorporates the aligned evidence: $\widetilde{\mathbf{f}}_A = \mathbf{f}_A + \mathcal{Z} \left( [\mathbf{f}_A,\overline{\mathbf{f}}_A] \right).$

We perform the alignment at multiple feature depths and concatenate the resulting aligned features with the corresponding VGGT representations to obtain the shared representation $\mathbf{z}_i$. This representation combines explicit metric correspondence with global multi-view context without constructing a depth-hypothesis cost volume.

\subsection{Geometry-Aware Gaussian Prediction}
\label{sec:geoforward_gaussians}

The aligned representation $\mathbf{z}_i$ is used to predict a Gaussian for each valid source pixel. Following the coordinate-query formulation and recent continuous depth prediction methods~\citep{yu2026infinidepth}, we use separate depth and attribute branches. The separation allows the metric geometry of a Gaussian center to be predicted independently from its appearance and spatial attributes, while both branches receive the same geometry-guided cross-view evidence.

\noindent\textbf{Metric depth and Gaussian centers.}
For pixel $i$, the depth branch combines the shared aligned representation with local RGB features to obtain a visual descriptor $\mathbf{v}_i$. Sparse LiDAR is additionally encoded as a metric prior $\boldsymbol{\ell}_i$ with support $s_i$. The visual and LiDAR-conditioned predictions are combined as
\begin{equation}
d_i = d_{\min} + (d_{\max}-d_{\min}) \sigma \left( \Psi_{\mathrm{vis}}(\mathbf{v}_i) + \Psi_{\mathrm{prior}} (\mathbf{v}_i,\boldsymbol{\ell}_i,s_i) \right),
\label{eq:gf_metric_depth}
\end{equation}
where $\sigma(\cdot)$ is the sigmoid function and $[d_{\min},d_{\max}]$ is the prescribed metric depth range. The LiDAR prior provides sparse metric information complementary to the correspondence mechanism rather than directly replacing the predicted depth.

The Gaussian center is obtained by camera unprojection: $\boldsymbol{\mu}_i = R_{t,v} \left( d_i K_{t,v}^{-1} \widetilde{\mathbf{p}}_i \right) + \mathbf{t}_{t,v}.$

We further derive camera-space surface normals $\mathbf{n}_i$ from the predicted depth and estimate a confidence value $c_i$ from detached visual features. These quantities provide geometric and reliability cues to the attribute branch.

\noindent\textbf{Gaussian attributes.}
The attribute branch predicts the remaining Gaussian parameters using the aligned representation, local RGB features, sparse-depth features, and source color. Its geometric conditioning descriptor is $\boldsymbol{\gamma}_i = [ \log d_i, \mathbf{n}_i, c_i, \nu_i ],$ where $\nu_i$ denotes camera validity. The resulting feature is $\mathbf{h}_i = \mathcal{D}_{\mathrm{attr}} \left( \mathbf{z}_i, \boldsymbol{\ell}_i, \boldsymbol{\gamma}_i, I_i \right),$ and the initial Gaussian attributes are predicted as $\mathbf{b}_i = \mathcal{A}(\mathbf{h}_i),$ including rotation, scale, opacity, and appearance parameters.
The shared representation $\mathbf{z}_i$ therefore allows the metric correspondence established in Sec.~\ref{sec:geoforward_alignment} to influence both Gaussian center locations through depth prediction and the attributes describing the local surface.

\subsection{Sparse 3D Attribute Refinement}
\label{sec:geoforward_refinement}

Although geometry-guided alignment improves the consistency of independently decoded Gaussians, neighboring Gaussians may still exhibit incompatible attributes after prediction. We therefore perform a lightweight refinement directly in 3D space while keeping the predicted centers fixed.

Specifically, Gaussian centers are grouped into sparse voxels according to their reference-frame coordinates. For the Gaussians $\mathcal{I}_j$ in voxel $j$, we first pool their latent features: $\overline{\mathbf{h}}_j = \frac{1}{|\mathcal{I}_j|} \sum_{i\in\mathcal{I}_j} \mathbf{h}_i.$ A sparse convolution $\mathcal{C}$ aggregates information from neighboring occupied voxels, and a readout $\mathcal{R}$ predicts an additive correction:
\begin{equation}
\mathbf{b}'_i = \mathbf{b}_i + \mathcal{R} \left( \mathcal{C} (\overline{\mathbf{h}}) \right)_{j(i)},
\label{eq:gf_attribute_refinement}
\end{equation}
where $j(i)$ denotes the voxel containing Gaussian $i$. Gaussians within the same voxel share the predicted correction while retaining their individual decoded attributes.

Importantly, the refinement does not modify $\boldsymbol{\mu}_i$. It therefore regularizes local Gaussian attributes without overriding the metric geometry established by the depth branch. The corrected parameters determine the covariance $\boldsymbol{\Sigma}_i$, opacity $o_i$, and appearance coefficients $\mathbf{a}_i$ after the corresponding activation functions. We express the base color as a bounded residual around the source pixel color and transform the covariance and appearance parameters into the common reference frame. Additional activation and initialization details are provided in the appendix.

\subsection{Temporal Gaussian Composition}
\label{sec:geoforward_temporal}

GeoForward predicts Gaussian representations independently at two endpoint timestamps. To render an intermediate time $\tau$, vehicle-associated Gaussians are translated according to their estimated motion and jointly composited.

Following~\citep{yu2026recondrive}, a frozen SAM2 model~\citep{ravi2025sam} is prompted with projected vehicle boxes to obtain object masks. Source-pixel Gaussians within accepted masks are associated with the corresponding annotated vehicle velocity $\mathbf{v}^{\mathrm{motion}}_i$, while Gaussians outside the accepted masks are assigned zero velocity. A Gaussian originating at timestamp $t_i$ is translated as
\begin{equation}
\boldsymbol{\mu}_i(\tau) = \boldsymbol{\mu}_i + (\tau-t_i) \mathbf{v}^{\mathrm{motion}}_i.
\label{eq:gf_motion}
\end{equation}
This temporal composition is applied after Gaussian reconstruction and does not modify the proposed cross-camera alignment mechanism.

\subsection{Learning Objectives}
\label{sec:geoforward_training}

We train GeoForward in two stages. The first stage learns endpoint geometry and appearance from source-view reconstruction, while the second stage introduces temporal composition and intermediate-view supervision. The VGGT encoder remains frozen throughout both stages.

\noindent\textbf{Endpoint reconstruction.}
For each endpoint, the predicted Gaussians are rendered into their source views and supervised using image and metric-depth observations. Offline dense metric-depth maps provide training targets only and are not used as model inputs. The endpoint objective is
\begin{equation}
\mathcal{L}_{\mathrm{end}} = \lambda_{\mathrm{img}} \mathcal{L}_{\mathrm{img}} + \lambda_{\mathrm{d}} \mathcal{L}_{\mathrm{d}} + \lambda_{\mathrm{conf}} \mathcal{L}_{\mathrm{conf}} + \lambda_{\mathrm{geo}} \mathcal{L}_{\mathrm{geo}} + \mathcal{L}_{\mathrm{reg}},
\label{eq:gf_single_loss}
\end{equation}
where $\mathcal{L}_{\mathrm{img}}$ measures rendered-image reconstruction, $\mathcal{L}_{\mathrm{d}}$ supervises metric depth, and $\mathcal{L}_{\mathrm{conf}}$ supervises the predicted confidence. $\mathcal{L}_{\mathrm{geo}}$ contains depth-boundary and multi-scale image constraints, while $\mathcal{L}_{\mathrm{reg}}$ regularizes Gaussian attributes. Detailed formulations are provided in the appendix.

Confidence targets are constructed from the discrepancy between predicted depth and the offline dense metric-depth targets, with the corresponding validity masks. Thus, confidence represents the reliability of the predicted geometry rather than an additional source of metric supervision.

\noindent\textbf{Temporal supervision.}
In the second stage, endpoint and intermediate renderings are jointly supervised. In addition to direct image reconstruction, we use a pose-based reprojection constraint to encourage temporal consistency. Within each camera, the first endpoint depth and known camera poses warp an intermediate observation into the endpoint coordinate system. On static and projection-valid regions, brightness-normalized $L_1$ and SSIM residuals define $\mathcal{L}_{\mathrm{proj}} = \mathcal{L}_{\mathrm{photo}} + \lambda_{\mathrm{ssim}} \mathcal{L}_{\mathrm{ssim}}.$ The temporal objective is $\mathcal{L}_{\mathrm{temporal}} = \lambda_{\mathrm{img}} \mathcal{L}_{\mathrm{img}} + \lambda_{\mathrm{proj}} \mathcal{L}_{\mathrm{proj}} + \mathcal{L}_{\mathrm{reg}}.$
The overall training procedure optimizes Gaussian reconstruction while using sparse LiDAR in two complementary ways: it establishes explicit cross-camera correspondence through the alignment module and provides sparse metric information to the depth prediction branch. Dense metric-depth targets are used only during training and do not replace the sparse geometric input available at inference time.
\par\endgroup

\section{Experiments}

\begin{table*}[t]
\centering
\caption{
Comparison of geometric accuracy and consistency
on nuScenes.
}
\label{tab:nuscenes_geometry_results}
\scriptsize
\setlength{\tabcolsep}{7pt}
\renewcommand{\arraystretch}{0.95}

\begin{tabular}{lcccccccc}
\toprule
Method
& Acc. $\downarrow$
& Comple. $\downarrow$
& CD $\downarrow$
& AbsRel $\downarrow$
& SqRel $\downarrow$
& RMSE $\downarrow$
& $\delta_{1.25}$ $\uparrow$
& DepthCons $\downarrow$ \\
\midrule

Street Gaussians~\cite{yan2024street}
& 2.505 & 7.136 & 9.641
& 0.719 & 10.081 & 15.450
& 0.100 & 5.992 \\

PVG~\cite{chen2026periodic}
& 9.033 & 3.946 & 12.979
& 0.780 & 14.951 & 8.210
& 0.589 & 4.538 \\

Deformable 3DGS~\cite{yang2024deformable}
& 2.927 & 6.420 & 9.347
& 0.818 & 24.110 & 14.407
& 0.250 & 8.196 \\

OmniRe~\cite{chen2025omnire}
& \underline{1.137}
& \textbf{0.489}
& \underline{1.625}
& \underline{0.152}
& \underline{1.898}
& \underline{5.229}
& \underline{0.833}
& 4.629 \\

DrivingForward~\cite{tian2025drivingforward}
& 1.228 & 1.195 & 2.423
& 0.255 & 3.365 & 7.900
& 0.666 & \underline{3.756} \\

ReconDrive~\cite{yu2026recondrive}
& 3.735 & 1.057 & 4.793
& 0.607 & 23.662 & 10.466
& 0.680 & 5.733 \\

DGGT~\cite{chen2026dggt}
& 1.690 & 5.459 & 7.149
& 0.421 & 6.041 & 10.477
& 0.461 & 4.985 \\

\textbf{GeoForward (Ours)}
& \textbf{0.674}
& \underline{0.761}
& \textbf{1.435}
& \textbf{0.148}
& \textbf{1.548}
& \textbf{5.016}
& \textbf{0.849}
& \textbf{2.446} \\

\bottomrule
\end{tabular}
\end{table*}

\begin{table*}[t]
\centering
\caption{
Comparison of geometric accuracy and consistency on Waymo.
}
\label{tab:waymo_geometry_results}

\scriptsize
\setlength{\tabcolsep}{7pt}
\renewcommand{\arraystretch}{0.95}

\begin{tabular}{lcccccccc}
\toprule
Method
& Acc. $\downarrow$
& Comple. $\downarrow$
& CD $\downarrow$
& AbsRel $\downarrow$
& SqRel $\downarrow$
& RMSE $\downarrow$
& $\delta_{1.25}$ $\uparrow$
& DepthCons $\downarrow$ \\
\midrule

Street Gaussians~\cite{yan2024street}
& 3.189 & 1.520 & 4.710
& 0.356 & 8.818 & 9.549 & 0.586 & 2.799 \\

PVG~\cite{chen2026periodic}
& 7.969 & 5.089 & 13.058
& 0.562 & 11.490 & 8.519 & 0.609 & 2.026 \\

Deformable 3DGS~\cite{yang2024deformable}
& 2.389 & 4.624 & 7.012
& 0.512 & 8.208 & 13.837 & 0.294 & 1.503 \\

STORM~\cite{yang2025storm}
& \underline{1.018}
& \underline{0.815}
& \underline{1.833}
& \underline{0.185}
& \underline{2.526}
& \underline{6.564}
& \underline{0.806}
& \textbf{1.272} \\

StreetForward~\cite{yu2026streetforward}
& 4.882 & 4.270 & 9.152
& 0.457 & 9.152 & 12.069 & 0.498 & 16.826 \\

Envision4D~\cite{song2026envision4d}
& 3.134 & 4.512 & 7.646
& 0.319 & 5.915 & 9.356 & 0.664 & 7.627 \\

\textbf{GeoForward}
& \textbf{0.569}
& \textbf{0.688}
& \textbf{1.257}
& \textbf{0.104}
& \textbf{0.875}
& \textbf{3.993}
& \textbf{0.917}
& \underline{1.477} \\

\bottomrule
\end{tabular}
\end{table*}

\begin{table*}[t]
\centering
\caption{Comparison of rendering quality on nuScenes and Waymo.}
\label{tab:nuscenes_waymo_results}
\scriptsize
\setlength{\tabcolsep}{7pt}

\resizebox{\textwidth}{!}{%
\begin{tabular}{@{}c@{\hspace{8pt}}c@{}}

{%
\renewcommand{\arraystretch}{1.20}
\begin{tabular}[t]{@{}lcccccc@{}}
\multicolumn{7}{c}{\textbf{nuScenes}} \\
\addlinespace[2pt]
\toprule
\multirow[c]{2}{*}{Method}
& \multicolumn{3}{c}{\shortstack{Visual Scene\\Reconstruction}}
& \multicolumn{3}{c}{\shortstack{Novel-View\\Synthesis}} \\
\cmidrule{2-4}
\cmidrule{5-7}
& PSNR $\uparrow$ & SSIM $\uparrow$ & LPIPS $\downarrow$
& PSNR $\uparrow$ & SSIM $\uparrow$ & LPIPS $\downarrow$ \\
\midrule

Street Gaussians~\cite{yan2024street}
& 26.090 & 0.774 & 0.430
& 18.942 & 0.621 & 0.515 \\

PVG~\cite{chen2026periodic}
& 33.040 & 0.923 & 0.213
& \textbf{27.416} & \textbf{0.810} & 0.302 \\

DeformableGS~\cite{yang2024deformable}
& 25.185 & 0.760 & 0.412
& 22.188 & 0.692 & 0.451 \\

OmniRe~\cite{chen2025omnire}
& 30.597 & 0.921 & 0.193
& 23.474 & 0.688 & 0.332 \\

DrivingForward~\cite{tian2025drivingforward}
& 23.791 & 0.878 & 0.131
& 18.240 & 0.665 & 0.303 \\

ReconDrive~\cite{yu2026recondrive}
& \underline{33.295} & \underline{0.961} & \underline{0.059}
& \underline{23.986} & 0.730 & \textbf{0.259} \\

DGGT~\cite{chen2026dggt}
& 22.503 & 0.803 & 0.305
& 22.498 & \underline{0.808} & 0.307 \\

\textbf{GeoForward (Ours)}
& \textbf{36.615} & \textbf{0.987} & \textbf{0.032}
& 23.191 & 0.730 & \underline{0.279} \\

\bottomrule
\end{tabular}
}

&

{%
\renewcommand{\arraystretch}{1.33}
\begin{tabular}[t]{@{}lcccccc@{}}
\multicolumn{7}{c}{\textbf{Waymo}} \\
\addlinespace[2pt]
\toprule
\multirow[c]{2}{*}{Method}
& \multicolumn{3}{c}{\shortstack{Visual Scene\\Reconstruction}}
& \multicolumn{3}{c}{\shortstack{Novel-View\\Synthesis}} \\
\cmidrule{2-4}
\cmidrule{5-7}
& PSNR $\uparrow$ & SSIM $\uparrow$ & LPIPS $\downarrow$
& PSNR $\uparrow$ & SSIM $\uparrow$ & LPIPS $\downarrow$ \\
\midrule

Street Gaussians~\cite{yan2024street}
& \underline{34.5317} & \textbf{0.9558} & 0.0985
& \underline{25.6590} & 0.7932 & 0.2333 \\

PVG~\cite{chen2026periodic}
& \textbf{35.5709} & 0.9505 & 0.1212
& \textbf{26.9835} & \underline{0.7998} & 0.2516 \\

DeformableGS~\cite{yang2024deformable}
& 27.3924 & 0.8327 & 0.3274
& 19.8208 & 0.6105 & 0.4921 \\

STORM~\cite{yang2025storm}
& 21.6821 & 0.7362 & 0.3025
& 21.0298 & 0.7032 & 0.3251 \\

StreetForward~\cite{yu2026streetforward}
& 25.9594 & 0.7872 & 0.1734
& 24.5113 & 0.7470 & \underline{0.2165} \\

Envision4D~\cite{song2026envision4d}
& 32.5785 & \underline{0.9518} & \textbf{0.0694}
& 25.0997 & 0.7490 & 0.2337 \\

\textbf{GeoForward (Ours)}
& 32.2823 & 0.9329 & \underline{0.0797}
& 25.5167 & \textbf{0.8026} & \textbf{0.1965} \\

\bottomrule
\end{tabular}
}

\end{tabular}%
}
\end{table*}

\begin{figure*}[t]
\captionsetup[subfigure]{labelformat=simple,labelsep=period}
\centering

\begin{subfigure}{\textwidth}
\centering
\includegraphics[
  scale=0.135,
  trim=0 -2 0 0,
  clip
]{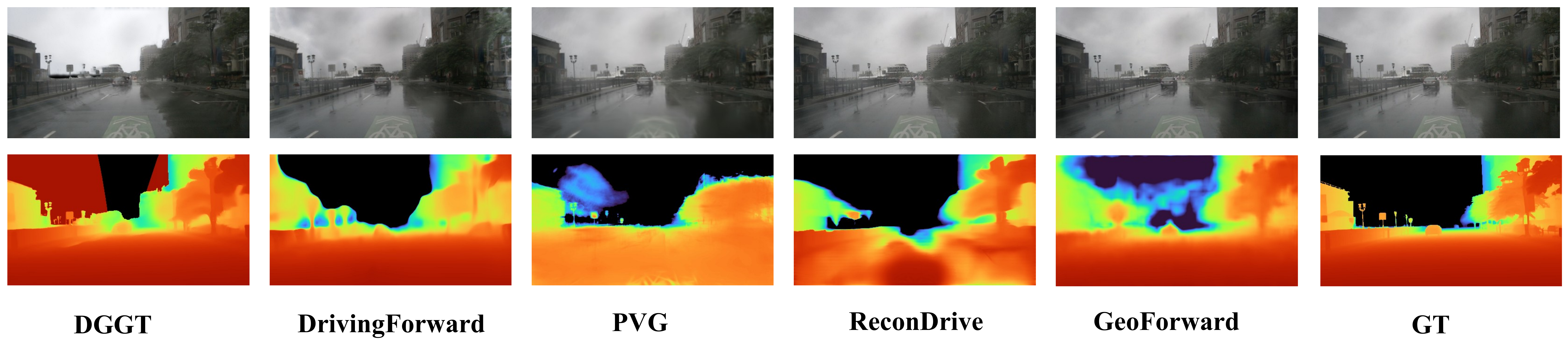}
\caption{Qualitative comparison in a nuScenes driving scene.}
\label{fig:exp1}
\end{subfigure}

\vspace{0.5em}

\begin{subfigure}{\textwidth}
\centering
\includegraphics[
  scale=0.135,
  trim=0 -2 0 0,
  clip
]{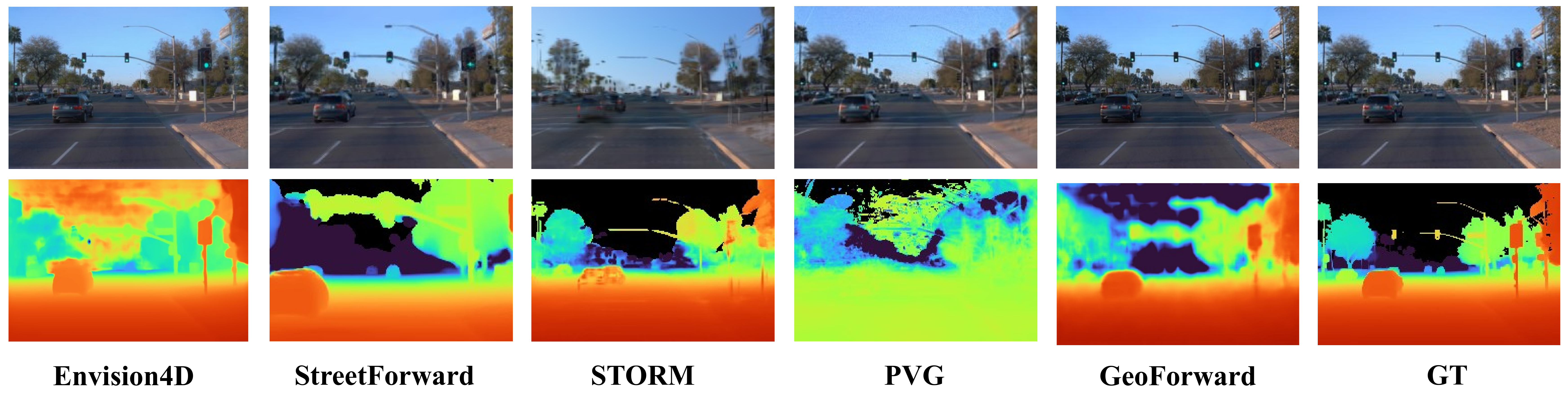}
\caption{Qualitative comparison in a dynamic traffic scene from the Waymo dataset.}
\label{fig:exp2}
\end{subfigure}

\caption{\textbf{Qualitative reconstruction results.} For each method, the upper row shows visual-view RGB renderings, and the lower row shows the corresponding depth maps. GT denotes ground truth.}
\label{fig:experiments}
\end{figure*}

\begin{table*}[t]
\centering

\begin{minipage}[t]{0.47\textwidth}
\vspace{0pt}
\centering
\caption{Component ablations on nuScenes: novel-view
rendering and depth quality.}
\label{tab:ablation_geometry}

\vspace{1mm}
\scriptsize
\renewcommand{\arraystretch}{0.95}

\resizebox{\linewidth}{!}{%
\begin{tabular}{@{}lccccc@{}}
\toprule
Setting
& PSNR $\uparrow$
& SSIM $\uparrow$
& LPIPS $\downarrow$
& RMSE $\downarrow$
& $\delta_{1.25}$ $\uparrow$ \\
\midrule

w/o LiDAR Align.
& 22.1891 & 0.6789 & 0.3188 & 7.0338 & 0.2765 \\

w/o Gaussian Head
& 20.0539 & 0.5754 & 0.4451 & 5.5165 & 0.8317 \\

w/o Motion Compensation
& 23.1392 & 0.7292 & 0.2796 & 5.0284 & 0.8466 \\

\textbf{Full GeoForward}
& \textbf{23.1912}
& \textbf{0.7302}
& \textbf{0.2789}
& \textbf{5.0166}
& \textbf{0.8486} \\
\bottomrule
\end{tabular}%
}
\end{minipage}
\hfill
\begin{minipage}[t]{0.47\textwidth}
\vspace{0pt}
\centering
\caption{Zero-shot geometric comparison on KITTI and Argoverse 2.}
\label{tab:geometry_comparison}

\vspace{1mm}
\scriptsize
\renewcommand{\arraystretch}{0.95}

\resizebox{\linewidth}{!}{%
\begin{tabular}{@{}lcccccc@{}}
\toprule
& \multicolumn{3}{c}{\textbf{KITTI}}
& \multicolumn{3}{c}{\textbf{Argoverse 2}} \\
\cmidrule{2-4}
\cmidrule{5-7}

Method
& Acc. $\downarrow$
& AbsRel $\downarrow$
& RMSE $\downarrow$
& Acc. $\downarrow$
& AbsRel $\downarrow$
& RMSE $\downarrow$ \\
\midrule

DGGT
& 64.3272 & 1.0991 & 25.5043
& 5.3868 & 0.5605 & 20.4651 \\

STORM
& 1.3977 & 0.2856 & 8.8221
& 2.5556 & 0.6836 & 14.6333 \\

\textbf{GeoForward}
& \textbf{1.0749} & \textbf{0.2596} & \textbf{7.3550}
& \textbf{1.6104} & \textbf{0.2982} & \textbf{10.0448} \\

\bottomrule
\end{tabular}%
}
\end{minipage}

\end{table*}

\begin{figure*}[t]
\centering
\includegraphics[scale=0.68,trim=0 -2 0 0,clip]{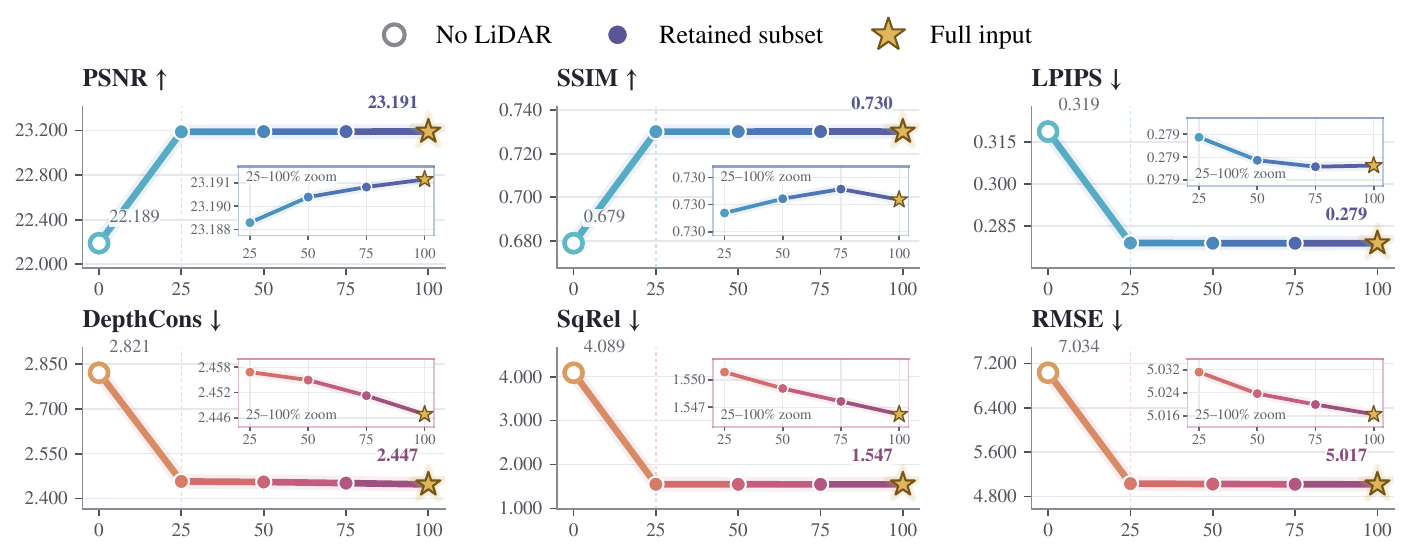}
\caption{\textbf{LiDAR retention ablation on nuScenes.} Performance largely saturates after retaining 25\% of the LiDAR points, with only minor changes at higher sampling densities.}
\label{fig:lidar_retention}
\end{figure*}

\noindent\textbf{Datasets.} We evaluate GeoForward on nuScenes~\cite{caesar2020nuscenes} and Waymo~\cite{sun2020scalability}, which provide calibrated multi-camera images and LiDAR measurements of driving scenes across varied weather and traffic conditions. For nuScenes, we train GeoForward on the official training split. For Waymo, we start from a nuScenes-trained checkpoint, fine-tune it on temporally sampled Waymo training windows, and then conduct the Waymo comparison. To assess transfer beyond these two benchmarks, we evaluate geometric reconstruction on KITTI~\cite{geiger2012we} and Argoverse~2~\cite{wilson2023argoverse} using known calibrations without target-dataset fine-tuning.

\noindent\textbf{Baselines.} We compare against per-scene optimization methods, including Street Gaussians~\cite{yan2024street}, PVG~\cite{chen2026periodic}, and Deformable 3DGS~\cite{yang2024deformable}. The feed-forward baselines include DrivingForward~\cite{tian2025drivingforward}, ReconDrive~\cite{yu2026recondrive}, and DGGT~\cite{chen2026dggt}. For Waymo, we also report STORM~\cite{yang2025storm}, StreetForward~\cite{yu2026streetforward}, and Envision4D~\cite{song2026envision4d}, which span distinct scene representation strategies.

\noindent\textbf{Evaluation metrics.} We evaluate scene reconstruction and novel-view synthesis using PSNR, SSIM, and LPIPS. For geometric quality, we report point-cloud accuracy, completeness, and Chamfer distance, as well as depth AbsRel, SqRel, RMSE, and $\delta_{1.25}$. DepthCons measures depth agreement across overlapping cameras after accounting for their relative calibration. Higher values are better for PSNR, SSIM, and $\delta_{1.25}$; lower values are better for the remaining metrics.\par

\subsection{Main Results}
\label{sec:main_results}

\noindent\textbf{Geometry.} We compare geometric accuracy and consistency in Tabs.~\ref{tab:nuscenes_geometry_results} and~\ref{tab:waymo_geometry_results}. On nuScenes, GeoForward ranks first on seven of eight metrics and second on completeness. It reduces CD from $1.625$ to $1.435$ and DepthCons from $3.756$ to $2.446$, improving over the next-best methods by $11.69\%$ and $34.88\%$. On Waymo, GeoForward again leads on seven of eight metrics. Its CD of $1.257$ and RMSE of $3.993$ improve over STORM by $31.42\%$ and $39.17\%$. Acc. decreases from $1.018$ to $0.569$, completeness from $0.815$ to $0.688$, and $\delta_{1.25}$ rises from $0.806$ to $0.917$. DepthCons ranks second at $1.477$, behind STORM. These results demonstrate improved surface placement and depth accuracy across both benchmarks, covering complementary aspects of reconstructed scene geometry.

\noindent\textbf{Appearance.} We evaluate scene reconstruction and novel-view synthesis in Tab.~\ref{tab:nuscenes_waymo_results}. On nuScenes, GeoForward achieves the best reconstruction PSNR of $36.615$\,dB, SSIM of $0.987$, and LPIPS of $0.032$. Compared with ReconDrive, it gains $3.320$\,dB in reconstruction PSNR and reduces LPIPS from $0.059$ to $0.032$. Its novel-view LPIPS of $0.279$ ranks second behind ReconDrive, while PVG leads novel-view PSNR and SSIM. On Waymo, GeoForward achieves the best novel-view SSIM of $0.8026$ and LPIPS of $0.1965$, reducing LPIPS from StreetForward's $0.2165$. Its novel-view PSNR of $25.5167$\,dB ranks third, and reconstruction LPIPS of $0.0797$ ranks second behind Envision4D. These results highlight its structural and perceptual quality at unseen viewpoints.

\noindent\textbf{Generalization.} We evaluate geometric transfer to KITTI and Argoverse~2 without target-dataset fine-tuning. In Tab.~\ref{tab:geometry_comparison}, GeoForward achieves the lowest accuracy error, AbsRel, and RMSE among the compared methods on both datasets. Its RMSE reaches $7.3550$ on KITTI and $10.0448$ on Argoverse~2, improving over STORM by $16.63\%$ and $31.36\%$. On Argoverse~2, Acc. decreases from STORM's $2.5556$ to $1.6104$, while AbsRel improves from the next-best DGGT result of $0.5605$ to $0.2982$. These gains support effective geometric transfer in both surface placement and metric depth across the evaluated sensor configurations, including different camera layouts and viewpoints.\par

\subsection{Ablation Studies}
\label{sec:ablation_studies}

\noindent Tab.~\ref{tab:ablation_geometry} separates the effects of geometric alignment, Gaussian prediction, and temporal composition.

\noindent\textbf{LiDAR-guided alignment.} Removing LiDAR-guided alignment lowers novel-view PSNR from $23.1912$ to $22.1891$\,dB and SSIM from $0.7302$ to $0.6789$, while depth RMSE rises from $5.0166$ to $7.0338$ and $\delta_{1.25}$ falls from $0.8486$ to $0.2765$. Thus, metric-guided correspondence improves both rendering quality and depth accuracy by reducing ambiguity in cross-view feature matching.

\noindent\textbf{Gaussian head.} Removing the Gaussian head causes the largest appearance loss: PSNR drops to $20.0539$\,dB, SSIM to $0.5754$, and LPIPS increases from $0.2789$ to $0.4451$. Its smaller effect on depth RMSE, from $5.0166$ to $5.5165$, suggests that this component is particularly important for converting the aligned representation into renderable Gaussian attributes with consistent appearance across camera views and improved visual fidelity in unseen camera viewpoints.

\noindent\textbf{Motion compensation.} Removing motion compensation changes novel-view PSNR by $0.0520$\,dB and RMSE by $0.0118$, although the vehicle boundaries in Fig.~\ref{fig:appendix_ablation} show more noticeable local differences that are diluted when metrics are averaged over large, predominantly static image regions.

\noindent\textbf{LiDAR sparsity.} Fig.~\ref{fig:lidar_retention} shows sharp, consistent improvements in all six reported metrics between $0\%$ and $25\%$ LiDAR retention. Additional points yield only small, occasionally non-monotonic changes, indicating that relatively sparse measurements already provide useful geometric anchors.\par

\section{Conclusion}
\label{sec:conclusion}

\noindent\textbf{Limitations.}
GeoForward requires calibrated multi-camera images and sparse LiDAR at inference. The ablation results show a substantial loss in depth accuracy when LiDAR-guided alignment is removed, which may limit its use when range measurements are unavailable. Cross-time feature projection also assumes static correspondences, while temporal rendering depends on tracked vehicle masks and velocity estimates. Untracked objects and non-rigid motion therefore remain challenging. Future work could improve robustness to missing geometric observations and extend temporal reconstruction beyond objects with reliable motion priors.

\noindent\textbf{Conclusion.}
We introduced GeoForward, which uses sparse LiDAR to align camera features for depth and Gaussian prediction from calibrated multi-camera observations of driving scenes. Bounded local refinement reduces residual cross-camera projection errors, while global multi-view context complements the aligned local features. This design incorporates explicit metric constraints into correspondence formation before predicting the Gaussian scene representation. GeoForward leads seven of eight geometric metrics on both nuScenes and Waymo, leads Waymo novel-view SSIM and LPIPS, and transfers to KITTI and Argoverse~2 without fine-tuning. These results show how metric-guided correspondence supports accurate and transferable driving-scene reconstruction.

\section*{AI Use Statement}
We used ChatGPT and Codex to assist with manuscript drafting and editing,
LaTeX formatting, literature searches, and interpretation and consistency
checking of experimental results. The authors take responsibility for
the final content, including all AI-assisted text, claims, and artifacts.

\bibliography{iclr2026_conference}
\bibliographystyle{plainnat}

\appendix

\section{Datasets}
\label{app:datasets}

\noindent\textbf{nuScenes.}
The nuScenes dataset~\cite{caesar2020nuscenes} contains 1,000 driving scenes of approximately 20 seconds each, with 700, 150, and 150 scenes in the official training, validation, and test splits. We use all six surround-view cameras and sparse LiDAR. Training uses the official training scenes, including 12\,Hz interpolated sequences. Each reconstruction window takes two endpoint observations six frames apart, spanning approximately 0.5\,s. Images are resized to $280\times518$ (height $\times$ width), with camera intrinsics scaled accordingly. The main comparison uses 14 validation scenes; the separate full-validation audit covers all 150 validation scenes.

\noindent\textbf{Waymo.}
The Waymo Open Dataset~\cite{sun2020scalability} provides calibrated camera images, LiDAR measurements, and vehicle poses. We use its five cameras at 10\,Hz and resize their images to $280\times518$, updating the intrinsics accordingly. The two endpoint observations are five frames apart, again spanning approximately 0.5\,s. The separate full-validation audit covers all 202 official validation contexts.

\noindent\textbf{Zero-shot datasets.}
We evaluate geometric generalization on Argoverse 2 and KITTI Tracking
without fine-tuning on either dataset. For Argoverse 2, all compared
methods use the same ten validation logs:
\par\smallskip\noindent
\resizebox{\linewidth}{!}{%
\begin{tabular}{@{}l@{\hspace{8pt}}l@{\hspace{8pt}}l@{}}
\texttt{05fa5048-f355-3274-b565-c0ddc547b315} &
\texttt{0b86f508-5df9-4a46-bc59-5b9536dbde9f} &
\texttt{185d3943-dd15-397a-8b2e-69cd86628fb7} \\
\texttt{25e5c600-36fe-3245-9cc0-40ef91620c22} &
\texttt{27be7d34-ecb4-377b-8477-ccfd7cf4d0bc} &
\texttt{280269f9-6111-311d-b351-ce9f63f88c81} \\
\texttt{2f2321d2-7912-3567-a789-25e46a145bda} &
\texttt{3bffdcff-c3a7-38b6-a0f2-64196d130958} &
\texttt{44adf4c4-6064-362f-94d3-323ed42cfda9} \\
\multicolumn{3}{c}{\texttt{5589de60-1727-3e3f-9423-33437fc5da4b}}
\end{tabular}%
}
\par\smallskip\noindent
For KITTI Tracking, we use the stereo cameras \texttt{image\_02} and
\texttt{image\_03}. All compared methods use the same three sequences
and frame ranges: \texttt{0001} (380--430), \texttt{0002} (140--220),
and \texttt{0006} (65--120).

\section{Baselines}
\label{app:baselines}
We briefly describe each method included in the nuScenes or Waymo comparison,
focusing on its scene representation and treatment of dynamics.

\noindent$\bullet$\enspace\textbf{Street Gaussians}~\cite{yan2024street} reconstructs the background
and foreground vehicles with separate Gaussian sets. Tracked vehicle poses
place the foreground Gaussians in the scene, and the method refines these
poses during scene-specific optimization. It also models time-dependent
appearance with four-dimensional spherical harmonics. This object-centric
baseline lets us compare feed-forward prediction with an optimized scene
whose moving vehicles are represented explicitly.

\noindent$\bullet$\enspace\textbf{PVG}~\cite{chen2026periodic} represents a dynamic urban scene
using Gaussians with periodic motion and time-varying visibility. Its
parameters describe vibration, lifespan, and the time at which opacity
peaks. PVG optimizes one temporal Gaussian representation per scene without
requiring annotated object boxes to assign motion to individual instances.
We include it to compare our reconstructed geometry and appearance with a
scene-optimized model of continuous temporal change.

\noindent$\bullet$\enspace\textbf{DeformableGS}~\cite{yang2024deformable} places Gaussians in a
canonical space and predicts timestamp-dependent changes to their properties
with a deformation network. These changes move the canonical representation
to the observed state of the scene. The method was introduced for monocular
dynamic reconstruction; its deformation formulation provides a comparison
for driving scenes with substantial camera motion and multiple viewpoints.

\noindent$\bullet$\enspace\textbf{OmniRe}~\cite{chen2025omnire} organizes an urban scene into a
Gaussian scene graph rather than assigning every actor the same motion
model. It reconstructs a static background, rigid actors, articulated
pedestrians, and other deformable objects with their corresponding
representations. The nodes are optimized jointly for each scene. OmniRe
therefore provides a strong comparison for visual fidelity and geometry when
instance structure and per-scene optimization are available.

\noindent$\bullet$\enspace\textbf{DrivingForward}~\cite{tian2025drivingforward} predicts a driving
scene's Gaussian representation from flexible surround-view input. It
couples depth and Gaussian prediction with image-based supervision from
neighboring spatial and temporal observations, allowing predictions from
separate cameras to be placed in a common scene. We compare with it to
examine the contribution of explicit metric-guided correspondence and
LiDAR-conditioned depth in a feed-forward setting.

\noindent$\bullet$\enspace\textbf{ReconDrive}~\cite{yu2026recondrive} builds on VGGT features and
uses distinct heads for Gaussian-center depth and the remaining Gaussian
attributes. It aligns endpoint Gaussians in time using object velocities
and combines them for dynamic rendering. ReconDrive is a closely related
feed-forward baseline for assessing projected cross-view alignment and
LiDAR-token conditioning in a depth-and-attribute prediction framework.

\noindent$\bullet$\enspace\textbf{DGGT}~\cite{chen2026dggt} reconstructs dynamic driving scenes
from sparse images without requiring input camera poses. It jointly
estimates camera parameters and per-frame Gaussian maps, while separate
dynamic predictions describe Gaussian motion and temporal visibility.
This baseline tests rendering and geometric quality when the model must
also infer camera geometry rather than relying on calibrated input poses.

\noindent$\bullet$\enspace\textbf{STORM}~\cite{yang2025storm} predicts dynamic 3D Gaussians and
their motion from sparse posed observations in a single forward pass.
Learned motion tokens define a set of velocity bases, and image-dependent
weights compose these bases into per-Gaussian motion. Transforming and
aggregating Gaussians across observed frames allows the method to supervise
motion through reconstruction. We include STORM on Waymo as a feed-forward
comparison with learned scene motion.

\noindent$\bullet$\enspace\textbf{StreetForward}~\cite{yu2026streetforward} reconstructs a dynamic
street scene without input poses or object tracks. Its temporal masked
attention extends VGGT-style feature aggregation to extract motion-aware
features from image sequences. The method predicts a unified Gaussian
representation for static and moving content together with pixelwise
velocities. We include it to compare with a feed-forward system that learns
motion without annotated tracks.

\noindent$\bullet$\enspace\textbf{Envision4D}~\cite{song2026envision4d} addresses future-view
extrapolation with a feed-forward 4D Gaussian representation. It predicts
future camera poses through iterative denoising and models nonlinear
scene evolution using temporal attention and conditioned motion lifting.
Progressive training stabilizes its future-frame predictions. Its Waymo
results provide a comparison to a model designed for future appearance and
motion prediction.

\section{Evaluation}
\label{app:evaluation}

We provide detailed definitions of the image reconstruction and geometric
metrics used in our experiments as follows.

\paragraph{Image Reconstruction Quality.}
Following ReconDrive~\cite{yu2026recondrive}, we report PSNR, SSIM, and LPIPS
for both visual scene reconstruction and novel-view synthesis. Let $\hat I$
and $I$ denote the rendered and reference images, respectively, and let $M$
be the total number of evaluated pixel values. These metrics are defined as
\[
\begin{aligned}
\mathrm{MSE}
&= \frac{1}{M}\sum_{j=1}^{M}(\hat I_j-I_j)^2,
\qquad
\mathrm{PSNR}
=10\log_{10}\left(\frac{L^2}{\mathrm{MSE}}\right), \\[2pt]
\mathrm{SSIM}(x,y)
&=
\frac{(2\mu_x\mu_y+C_1)(2\sigma_{xy}+C_2)}
     {(\mu_x^2+\mu_y^2+C_1)(\sigma_x^2+\sigma_y^2+C_2)}, \\[2pt]
\mathrm{LPIPS}(I,\hat I)
&=
\sum_l\frac{1}{H_lW_l}
\sum_{h,w}
\left\|
\mathbf{w}_l\odot
\left(
\tilde{\phi}_l(I)_{hw}
-
\tilde{\phi}_l(\hat I)_{hw}
\right)
\right\|_2^2 .
\end{aligned}
\]
Here, $L$ is the maximum image intensity. For SSIM, $x$ and $y$ denote
corresponding local image windows; $\mu$, $\sigma^2$, and $\sigma_{xy}$ are
their local means, variances, and covariance, while $C_1$ and $C_2$ are
stabilizing constants. For LPIPS, $\tilde{\phi}_l(\cdot)$ denotes normalized
deep features at layer $l$, and $\mathbf{w}_l$ denotes the corresponding
channel weights. PSNR measures pixel-level fidelity, SSIM evaluates luminance,
contrast, and structural similarity, and LPIPS measures perceptual discrepancy
in deep feature space. Higher PSNR and SSIM and lower LPIPS indicate better
image quality.

\paragraph{Point-Cloud Reconstruction.}
Following DVGT~\cite{zuo2026dvgt}, let $\mathcal{P}$ and $\mathcal{G}$ denote
the predicted and reference point clouds, respectively. Accuracy and
Completeness are defined as
\[
\begin{aligned}
\mathrm{Accuracy}
&=
\frac{1}{|\mathcal{P}|}
\sum_{\mathbf{p}\in\mathcal{P}}
\min_{\mathbf{g}\in\mathcal{G}}
\|\mathbf{p}-\mathbf{g}\|_2,
\qquad
\mathrm{Completeness}
=
\frac{1}{|\mathcal{G}|}
\sum_{\mathbf{g}\in\mathcal{G}}
\min_{\mathbf{p}\in\mathcal{P}}
\|\mathbf{g}-\mathbf{p}\|_2 .
\end{aligned}
\]
Accuracy measures the fidelity of the predicted geometry, whereas Completeness
measures its coverage of the reference geometry. Following
PGSR~\cite{chen2024pgsr}, the symmetric Chamfer Distance is
\[
\mathrm{CD}=\mathrm{Accuracy}+\mathrm{Completeness},
\]
which jointly accounts for reconstruction fidelity and coverage.

\paragraph{Depth Accuracy.}
Let $\hat d_i$ and $d_i$ denote the predicted and reference metric depths at
valid pixel $i$, respectively, and let $N$ be the number of valid pixels.
Following DVGT~\cite{zuo2026dvgt}, we use ray depth, defined as the Euclidean
distance from a reconstructed 3D point to the ego-vehicle center of its
corresponding frame. The depth metrics are
\[
\begin{alignedat}{2}
\mathrm{AbsRel}
&= \frac{1}{N}\sum_{i=1}^{N}
\frac{|\hat d_i-d_i|}{d_i},
\qquad&
\mathrm{SqRel}
&= \frac{1}{N}\sum_{i=1}^{N}
\frac{(\hat d_i-d_i)^2}{d_i}, \\[2pt]
\mathrm{RMSE}
&= \sqrt{\frac{1}{N}\sum_{i=1}^{N}(\hat d_i-d_i)^2},
\qquad&
\delta_{1.25}
&= \frac{1}{N}\sum_{i=1}^{N}
\mathbb{I}\!\left[
\max\!\left(
\frac{\hat d_i}{d_i},
\frac{d_i}{\hat d_i}
\right)<1.25
\right].
\end{alignedat}
\]
AbsRel measures average relative depth error. SqRel and RMSE assign greater
penalties to large depth deviations, with RMSE directly measuring their
magnitude in metric units. The threshold accuracy $\delta_{1.25}$ reports the
fraction of predictions lying within a multiplicative factor of $1.25$ of the
reference depth.

\paragraph{Cross-Camera Consistency.}
Following CylinderDepth~\cite{abualhanud2026cylinderdepth}, DepthCons measures
whether overlapping cameras predict consistent metric distances for the same
physical points. Correspondences are established using reference depth and
known metric relative poses, after which the corresponding predictions are
expressed in a common ego coordinate system. Let $r_i^a$ and $r_i^b$ denote
the predicted distances to the common origin from cameras $a$ and $b$.
DepthCons is defined as
\[
\mathrm{DepthCons}
=
\sqrt{
\frac{1}{N_c}
\sum_{i=1}^{N_c}
(r_i^a-r_i^b)^2
},
\]
where $N_c$ denotes the number of valid cross-camera correspondences.
DepthCons measures agreement between views rather than their individual
accuracy with respect to the reference geometry.

Lower Accuracy, Completeness, CD, AbsRel, SqRel, RMSE, and DepthCons indicate
better geometric quality, whereas a higher $\delta_{1.25}$ is preferred.

\section{Additional Results}
\begin{figure*}[h]
\centering
\includegraphics[scale=0.115,trim=0 -2 0 0,clip]{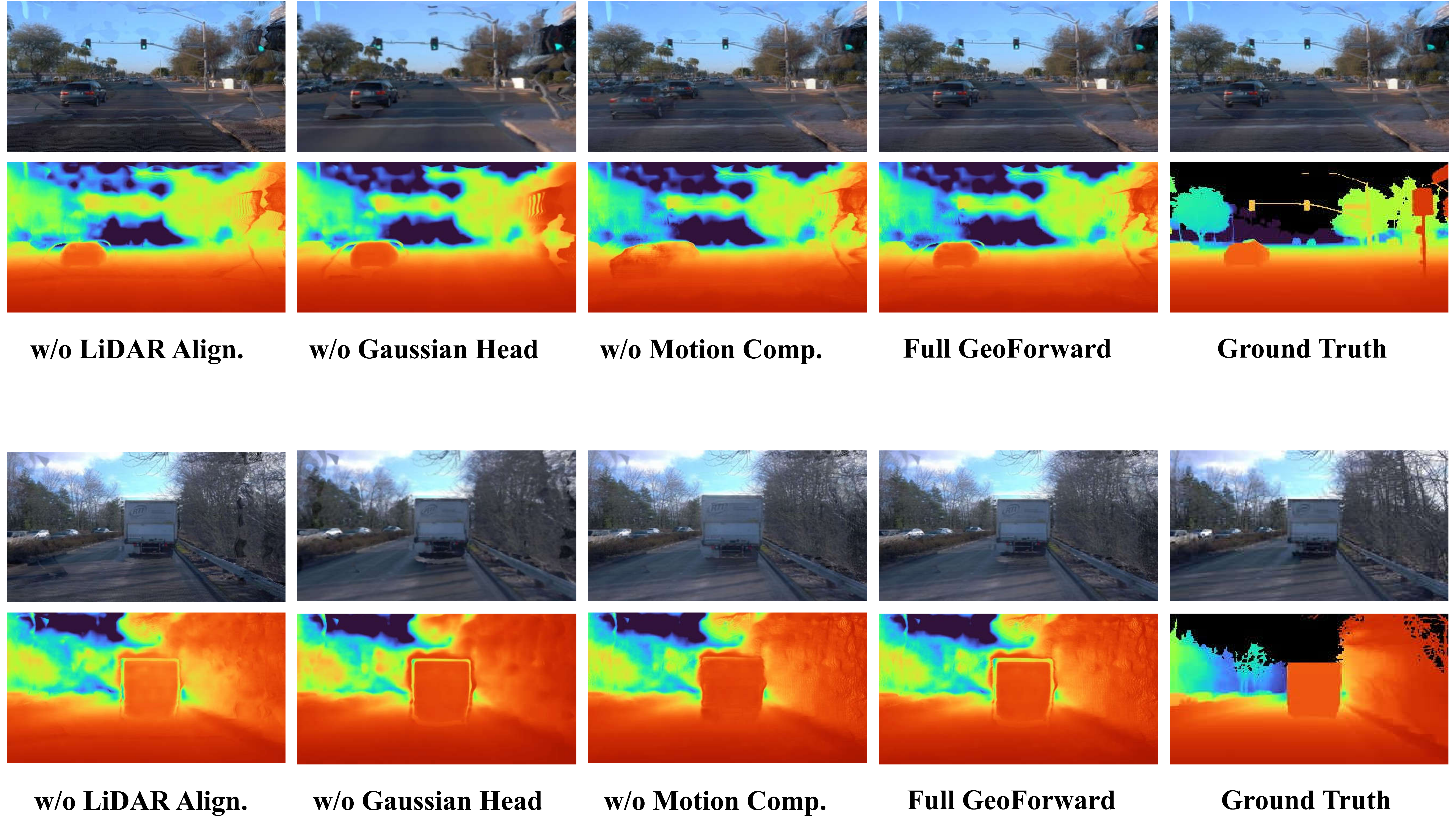}
\caption{\textbf{Qualitative ablation results on nuScenes.} For each of two scenes, the upper row shows RGB renderings and the lower row shows depth maps. We compare variants without LiDAR-guided alignment, the Gaussian head, or motion compensation with the full GeoForward model and ground truth (GT).}
\label{fig:appendix_ablation}
\end{figure*}

Figure~\ref{fig:appendix_ablation} shows an intersection with a foreground car and a tree-lined road with a truck. Removing LiDAR-guided alignment introduces visible horizontal artifacts on the nearby road and makes the curb and depth transitions less stable. Without the Gaussian head, the car, truck, and roadside details appear softer; the depth boundary around the truck also becomes broader. Without motion compensation, the car shows a displaced, ghosted outline, and its predicted depth extends beyond the compact vehicle silhouette in the full model. The truck boundary in the second scene likewise becomes less distinct. With all three components, GeoForward retains clearer vehicle contours and more continuous road surfaces in both RGB and depth. 

\begin{figure*}[h]
\centering
\includegraphics[scale=0.115,trim=0 -2 0 0,clip]{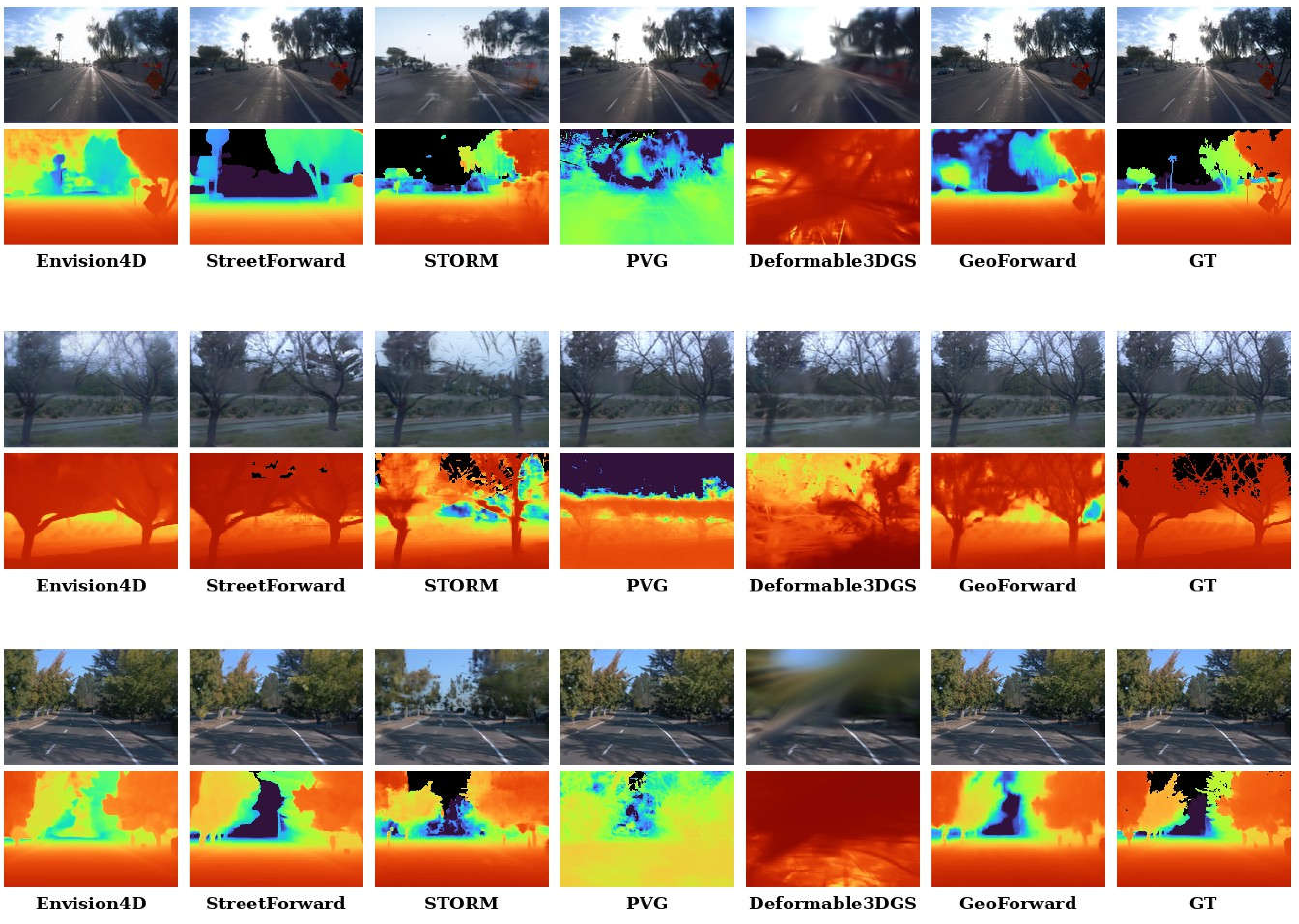}
\caption{\textbf{Qualitative comparison with baselines.} Three driving scenes are shown, each with RGB renderings above the corresponding depth maps. Columns compare Envision4D, StreetForward, STORM, PVG, DeformableGS, GeoForward, and ground truth (GT).}
\label{fig:appendix_baseline_comparison}
\end{figure*}

Figure~\ref{fig:appendix_baseline_comparison} compares a strongly backlit street view and two scenes with prominent roadside trees. Envision4D preserves much of the road appearance, but its depth maps smooth over tree structure, especially in the second scene. StreetForward renders recognizable RGB images, while its depth maps contain broad, nearly uniform regions around vegetation and miss thin branches. STORM loses detail in the backlit RGB view and produces fragmented depth around tree canopies. PVG retains lane markings and the overall RGB layout, but individual tree silhouettes are less distinct in depth; in the second scene, broad horizontal depth bands replace much of the tree structure. Deformable3DGS produces noticeably blurred RGB images in the first and third scenes, together with nearly uniform depth over large regions. GeoForward retains road markings and roadside objects while producing more distinguishable tree trunks, vegetation boundaries, and road regions in depth. Fine branches remain less detailed than in the ground truth, and regions without displayed ground-truth depth should not be interpreted as pixelwise depth agreement.

\begin{figure*}[t]
\centering
\includegraphics[scale=0.115,trim=0 -2 0 0,clip]{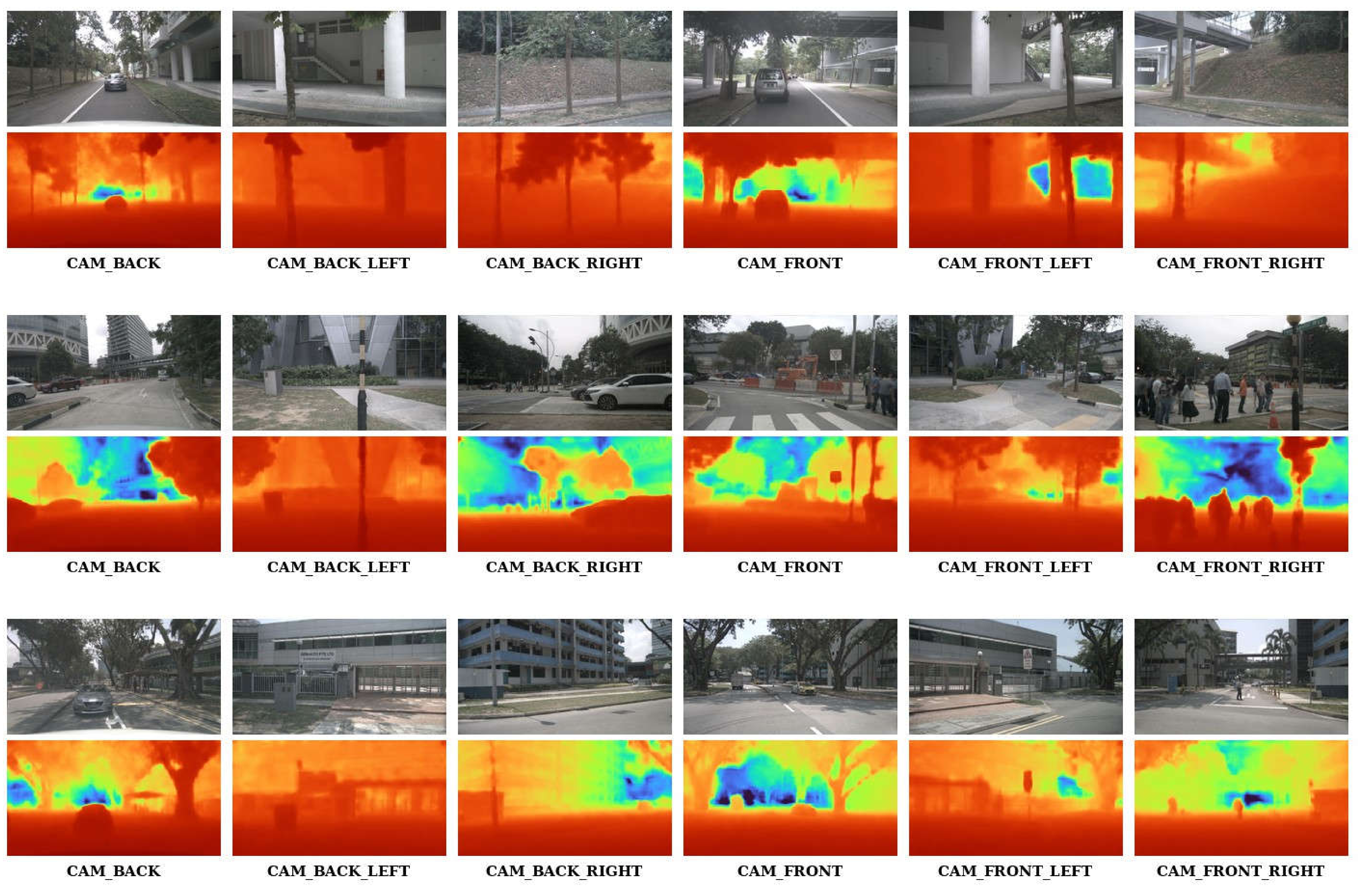}
\caption{\textbf{Six-camera qualitative results on nuScenes.} Each row shows one scene. From left to right, the columns correspond to CAM\_BACK, CAM\_BACK\_LEFT, CAM\_BACK\_RIGHT, CAM\_FRONT, CAM\_FRONT\_LEFT, and CAM\_FRONT\_RIGHT. Each camera view shows the rendered RGB image above its colorized predicted depth map.}
\label{fig:appendix_nuscenes_surround}
\end{figure*}

Figure~\ref{fig:appendix_nuscenes_surround} presents GeoForward's outputs from all six cameras across three nuScenes scenes. The views cover different road directions, building facades, vegetation, and lighting conditions. The RGB renderings retain recognizable scene structure in both forward-facing and rear-facing views. In the predicted depth maps, road surfaces form largely continuous regions, while vehicles, buildings, and trees introduce distinct depth boundaries. These examples complement the single-view comparisons by showing the reconstructed appearance and geometry around the vehicle.

\vspace{0.5cm}

\end{document}